\documentclass[fleqn,10pt]{wlscirep}
\usepackage[utf8]{inputenc}
\usepackage[T1]{fontenc}

\usepackage{color}
\definecolor{cardinal}{rgb}{0.77, 0.12, 0.23}
\definecolor{officegreen}{rgb}{0.0, 0.5, 0.0}
	\definecolor{lightbrown}{rgb}{0.71, 0.4, 0.11}

\title{On the Interpretability of Part-Prototype Based Classifiers: A Human Centric Analysis}

\author[1,*]{Omid Davoodi}
\author[2]{Shayan Mohammadizadehsamakosh}
\author[1]{Majid Komeili}
\affil[1]{Carleton University, School of Computer Science, Ottawa, ON, Canada}
\affil[2]{Sharif University of Technology, Department of Computer Engineering, Tehran, Iran}

\affil[*]{omid.davoudi@carleton.ca}

\begin{abstract}
Part-prototype networks have recently become methods of interest as an interpretable alternative to many of the current black-box image classifiers. However, the interpretability of these methods from the perspective of human users has not been sufficiently explored. In this work, we have devised a framework for evaluating the interpretability of part-prototype-based models from a human perspective. The proposed framework consists of three actionable metrics and experiments.
To demonstrate the usefulness of our framework, we performed an extensive set of experiments using Amazon Mechanical Turk. They not only show the capability of our framework in assessing the interpretability of various part-prototype-based models, but they also are, to the best of our knowledge, the most comprehensive work on evaluating such methods in a unified framework.

\end{abstract}
\begin{document}

\flushbottom
\maketitle
%
%
\thispagestyle{empty}


\section*{Introduction}

As Artificial Intelligence and Machine Learning have become more ubiquitous in many parts of the society and economy, the need for transparency, fairness, and trust increases. Many of the state-of-the-art methods and algorithms are black boxes where the decision-making process is opaque to humans. Interpretable and Explainable Artificial Intelligence aims to address this issue by offering methods that either explain the decisions of black-box models or are inherently interpretable themselves.

\begin{figure}[ht]
\centering
\includegraphics[width=\linewidth]{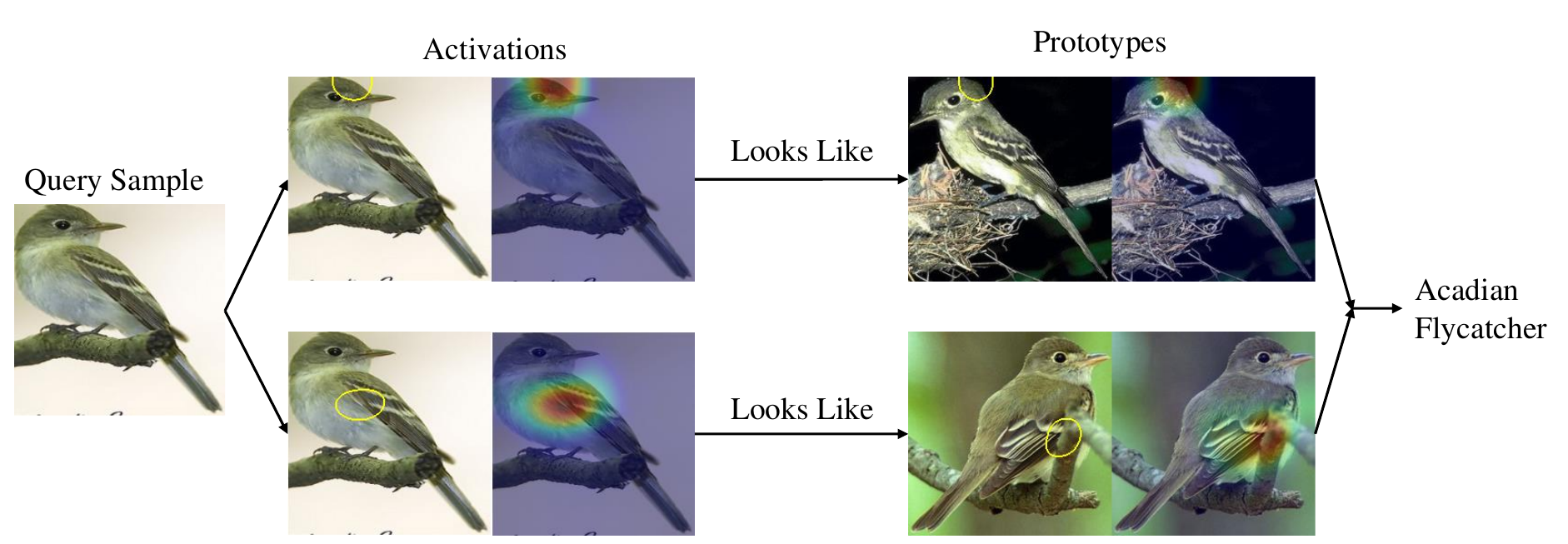}
\caption{Example of the decision-making process of a part-prototype method.}
\label{fig:howitworks}
\end{figure}

Prototype-based classifiers are a category of inherently interpretable methods that use prototypical examples to make their decisions. It is assumed that as long as the prototypes themselves are understandable by a human, the decision itself is interpretable\cite{molnar2020interpretable}. Prototype-based classifiers are not new inventions. Many existed long before the need for interpretability became so urgent\cite{bezdek1977prototype, kohonen1990improved, kuncheva1998nearest, seo2003soft, graf2009prototype}. In recent years, newer methods have been proposed that combine the power and expressablility of neural networks with the decision-making process of a prototype based classifier to create prototypical neural nets\cite{li2018deep, davoudi2021toward}, reaching results competitive with the state of the art while being inherently interpretable in the process.

A newer subcategory of prototype-based classifiers is part-prototype networks. These networks, usually operating in the domain of image classification use regions of a query sample, as opposed to the entire query image, to make their decisions. ProtoPNet\cite{chen2019looks} is the first of such methods that offered fine-grained explanations for image classification while offering state-of-the-art accuracy. Figure \ref{fig:howitworks} shows an example of how a part-prototype method makes its decisions.

The explanations given by these methods can be very different from each other. Even when the general layout of the explanation is similar, the part-prototypes themselves can be vastly different. It is unusual to assume that they offer the same level of interpretability. Therefore, the evaluation of their interpretability is necessary. 

While many of these methods evaluate the performance of their models and compare them to the state of the art, few analyze the interpretability of their methods. Most of the analysis in this regard seems to be focused on automatic metrics for assessing interpretability\cite{huangevaluation}. Such automatic metrics, while useful, are not a replacement for human evaluation of interpretability. Others have worked on human-assisted debugging \cite{bontempelli2022concept} but have not extended that to a full evaluation of method interpretability. 

Kim et al. offered a method for evaluating visual concepts by humans and even performed experiments on ProtoPNet and ProtoTree\cite{kim2022hive}, but their evaluation suffers from a number of issues. The scale of the experiments in Kim et al. is small, with only two part-prototype methods evaluated using only a single dataset. The experimental design of that work also relies on fine-grained ratings by human annotators. This type of design can be an unreliable way of measuring human opinion when there is no consensus on what each option means \cite{Krosnick2018}. It used the class label to measure the quality of the prototypes in the CUB dataset even though there was no indication that the human users were familiar with the minutiae of the distinctions between 200 classes of birds. Lastly, it used the default rectangular representation of prototypes from ProtoPNet and ProtoTree. These representations are prone to being overly broad and misleading to the human user compared to the actual activation heatmap. As a result, we propose a human-centric analysis consisting of a set of experiments to assess the interpretability of part-prototype methods.



\subsection*{Goals}

\begin{figure}[t]
\centering
\includegraphics[width=\linewidth]{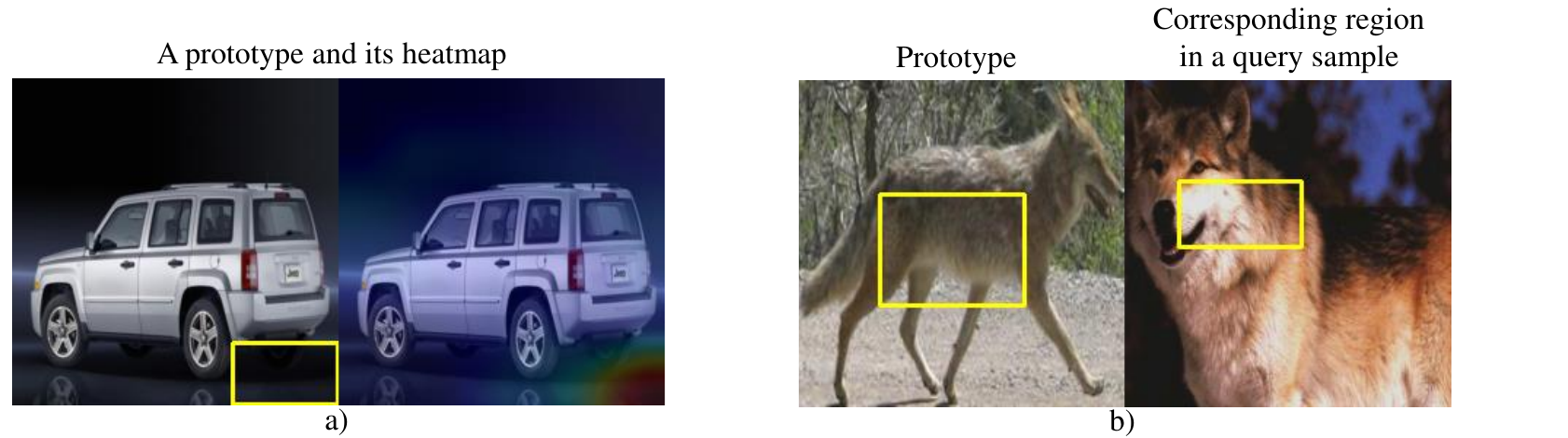}
\caption{Examples of interpretability problems with prototypes. a) The prototype itself is not interpretable because it is pointing to an irrelevant background region. b) lack of similarity between a prototype and the corresponding region in the query sample. }
\label{fig:badprot}
\end{figure}

The interpretability of a part-prototype system is not a well-defined concept. In this work, we focus on three properties that such systems should have in order to be interpretable.

\begin{itemize}
\item Interpretability of the prototype itself: The concept a prototype is referring to should be recognizable and understandable to a human. Figure \ref{fig:badprot} (a) shows an example of a prototype that is not interpretable because it points to an irrelevant background region. Machine learning methods and neural networks, in particular, can make correct decisions based on feature combinations in the data that a human might not understand. In addition, the presentation of such features is very important. A prototype might refer to a very unusual concept but its presentation might lead a human to wrongfully believe that they understand the reasoning behind a decision.

\item The similarity of a prototype to the corresponding region in the query sample: Even if the prototype itself is easily understood by a human, its activation on the query sample might not show the same concept as the prototype. Figure \ref{fig:badprot} (b) shows an example of this problem. This is important because it shows that the structural similarity in the embedding space that the prototypes reside in is not compatible with human understanding of similarity. This is a problem that has been reported in previous literature\cite{hoffmann2021looks}.

\item The interpretability of the decision-making process itself is also an important aspect of prototype-based methods. Even if the prototypes and their similarity to the activated patches of the query sample are understood by humans, the final decision might not be. For example, a model might select and use unrelated prototypes to correctly classify a sample.
\end{itemize}

The main novelty of this work is a more robust framework for evaluating the interpretability of part-prototype-based networks using human annotators. Some previous methods have tried to do such evaluations based on automatic metrics \cite{huangevaluation}, and some other works have worked on human-based evaluation of interpretability for other types of explainable AI methods \cite{lage2018human, colin2022cannot}. The closest work is HIVE \cite{kim2022hive} which suffers from a number of issues that are addressed in our approach. More on this will follow in the next section. 

Another novelty of this work is the proposal of three actionable metrics and experiments for evaluating the interpretability of part-prototype-based classifiers. We believe that if a model fails these tests, it would not be a good interpretable model. These can assist future researchers in providing evidence rather than just making assumptions about the interpretability of their approaches. 

Finally, our extensive set of experiments using Amazon Mechanical Turk includes comparisons of six related methods on three datasets. To the best of our knowledge, this is the most comprehensive work on evaluating the interpretability of such methods in a unified framework.

\section*{Background Information}

\subsection*{HIVE}

HIVE\cite{kim2022hive} (Human Interpretability of Visual Explanations) is a framework for human evaluation of interpretability in AI methods that use visual explanations. HIVE uses questionnaires to ask humans to rate multiple aspects of the interpretability of AI methods. These include similarity, agreement, and distinction tests.

There are, however, some problems with the question design of the HIVE framework that makes it flawed for the purpose of assessing interpretability for prototype-based methods. In particular, they ask the human participants to rate multiple criteria in a fine-grained manner. This practice has been shown to be unreliable because human opinion can differ greatly when asked to quantify vague concepts \cite{Krosnick2018}. HIVE also assumes that every bit of explanation is interpretable by itself, even though there are examples showing the opposite (see Figure \ref{fig:badprot}-a). Finally, HIVE does not follow the same process that is used by the AI model to classify a query sample. It splits the prototypes into class categories when showing them to human participants and hides the activation values. This is completely opposite of the way many of the part-prototype methods operate. Our proposed method does not suffer from these deficiencies.

\subsection*{ProtoPNet}
ProtoPNet\cite{chen2019looks} is a neural-network-based model that utilizes the similarity between parts of an input to a number of part-prototypes to classify images. The image first goes through a backbone network, usually a pre-trained image network such as ResNet consisting of multiple convolutional layers, and is turned into a spatially related matrix of embeddings. These embeddings are usually obtained in the penultimate layers of the common pre-trained networks. The distance between the closest of each embedding to each prototype is then fed into a fully connected output layer with softmax to determine the final class.

The resulting network is end-to-end trainable via gradient descent. A weighted sum of multiple loss functions is used to train the model. These losses include Cross-entropy loss for the final classification loss, as well as a Clustering loss designed to make sure all query samples have at least one prototype that is close to some part of them. There is also a Separation loss designed to make sure samples are far from prototypes of other classes.

\subsection*{SPARROW}
SPARROW\cite{kraft2021sparrow} is a method based on ProtoPNet that aims to make sure that the prototypes are semantically coherent. In principle, it aims to enforce sparsity, narrowness of scope and uniqueness to the prototypes. It does so by adding two additional components to the loss equation. The first one tries to decrease angular similarity between prototypes themselves to make sure they are unique, and the second one tries to make sure the prototypes themselves are as close as possible to at the query samples in the embedding space of the model in the hope of making the scope of the prototypes narrower.

\subsection*{Deformable ProtoPNet}
Deformable ProtoPNet\cite{donnelly2022deformable} is another method that aims to overcome some of the issues of the original. In this case, the rigidity of the prototypes in ProtoPNet. The prototypes in ProtoPNet are vectors with a spatially rigid semantic meaning in the pixel space. Deformable ProtoPNet aims to remedy that by creating spatially deformable prototypes where the position of features within each prototype can be different while still encoding the same concept.

\subsection*{ProtoTree}
By leveraging the interpretability of the decision trees, a new prototype-based model was introduced called the Neural Prototype Tree (ProtoTree)\cite{nauta2021neural}. A ProtoTree consists of a Convolutional Neural Network followed by a soft binary tree structure with learnable prototypes in its decision nodes. The distance between the nearest latent patch and a prototype determines to what extent the prototype is present anywhere in an input image, which influences the routing of the image through the corresponding node. As it uses a soft decision tree, a query sample is routed through both children, each with a certain weight. After that, the model makes its prediction based on the taken paths to all leaves and their class distributions. However, the authors claim that a trained soft ProtoTree can convert into a hard ProtoTree without losing accuracy. In contrast with the ProtoPNet with a linear bag-of-prototypes, the ProtoTree enforces a sequence of steps and supports negative associations. Consequently, it can drastically reduce the number of prototypes.

\subsection*{TesNet}
Another interpretable network architecture was designed by introducing a plug-in transparent embedding space (TesNet)\cite{wang2021interpretable} with basis concepts constructed on the Grassmann manifold. Concerning ProtoPNet, the authors mention two main disadvantages. First, the prototypes are implicitly assumed to follow a Gaussian distribution, which is improper for the complex data structure. Moreover, it cannot explicitly ensure that the learned prototypes are disentangled, an essential property for interpretable learning. The TesNet model was designed to alleviate such problems by proposing a transparent embedding space. In terms of architecture, the prototype layer in the ProtoPNet is replaced by a Transparent Subspace layer with "basis concepts". This layer covers as many subspaces as the number of classes, and each subspace is spanned by M basis concepts. The M within-class concepts are assumed orthogonal to each other, and each concept can be traced back to the high-level patches of the feature map. To this end, it defines rigorous losses. They include orthogonality of within-class concepts, subspace separation, high-level patche grouping, and concept-based classification loss.

\subsection*{ProtoPool}
After ProtoTree and TesNet, ProtoPool\cite{rymarczyk2022interpretable} was introduced. The model is said to have mechanisms that reduce the number of prototypes and obtain higher interpretability and easier training. The first is achieved through prototype sharing, the second comes from defining a novel similarity function, and the third results from employing a soft assignment based on prototype distributions to optimally use prototypes from the pool. The prototypes must focus on a salient visual feature, not the background. For this purpose, instead of maximizing the global activation, they widen the gap between the maximal and average similarity between the image activation map and prototypes. The other two advantages come from having a pool of prototypes and learning distributions of them per class. Therefore, the architecture is very similar to ProtoPNet, with some changes in the prototype pool layer, which contribute to prototype distribution and applying the new focal similarity function.

\subsection*{ACE}
ACE\cite{ghorbani2019towards} is an unsupervised method for extracting visual concepts from a dataset. These concepts are small regions of the images that seem to be important identifiers of the subjects of the pictures. To find these concepts, first, the images are segmented into small pieces, and then each piece is fed to a state-of-the-art classifier network. The output of the bottleneck layer of this network is then used as a representation of the concept present in the segment. Clustering methods are then utilized to find important clusters within these representations which are then picked as important concepts for the dataset.

\subsection*{Datasets Used}
We used three datasets in our experiments. The datasets used had to deal with concepts that were broadly understandable to the average person. Note that it is not necessary for a human to be able to distinguish between the classes within a dataset (for example, distinguishing between different species of birds). The only requirement is for people to understand the general concept of the subjects of the image (For example, knowing what a bird is). 

\subsubsection*{CUB-200-2011}
CUB-200-2011\cite{WahCUB_200_2011} is a dataset of birds consisting of 11988 total images that are split into 200 classes. This is a popular dataset in part-prototype network research. It also depicts subjects that the average human can understand. For our purposes, we split the dataset into two halves with 6194 images for training and 5994 images for testing. The reason for such a drastic split is that the augmentation method we used (which is the same as the one used for ProtoPNet and many of the other derived methods) multiplies the number of training images by a factor of 30.

\subsubsection*{Stanford Cars Dataset}
The Stanford Cars Dataset\cite{6755945} is an image dataset consisting of 196 different models of cars. They are split into 8339 training and 8237 test images. This is another dataset where the subjects are generally well understood by an average human and thus, was a good pick for our experiments. The same augmentation process was also done on this dataset except that there are no bounding box coordinates available for this dataset. As a result, the images were not cropped to the subjects.

\subsubsection*{ImageNet Subset}
ImageNet\cite{deng2009imagenet} is a very diverse image dataset containing a hierarchy of classes. We chose 7 classes out of these and created a subset that we believed was understandable by the average human while also containing similar concepts and a relatively balanced number of samples per class. These classes were "Big Cats", "Collie breeds of dogs", "Domestic Cats", "Foxes", "Swiss Mountain Dogs", "Wild Cats" and "Wolves". It contains 16586 training and 1686 test images. The same augmentation process was also done on this dataset.

\section*{Methodology}

\begin{figure}[t]
\centering
\includegraphics[width=\linewidth]{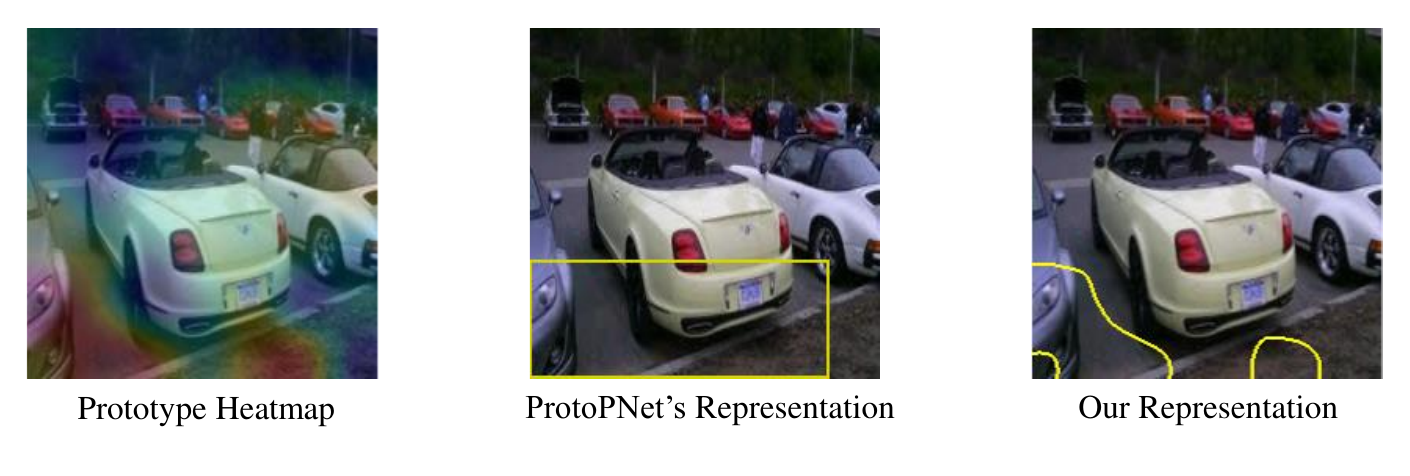}
\caption{Examples of complex prototype heatmaps and the failure of a rectangular prototype representation. While the actual heatmap (left) points to irrelevant background regions, the rectangular representation (center) misleadingly includes foreground regions. Our solution (right) properly represents the actual heatmap.  }
\label{fig:badrepr}
\end{figure}

The interpretability of a prototype-based machine learning method as a whole is too broad to be effectively evaluated. As a result, our experiments were designed to measure the human opinion on the individual properties required for an interpretable method. This focus on single aspects allowed us to gain fine-grained insight into how much human users understand and agree with the explanations of different methods.

Human annotators were recruited via Amazon Mechanical Turk. To ensure that workers have a good understanding of the task, they were required to first complete a qualification test. Furthermore, to ensure the quality of the responses and reduce noise, we inserted a few validation samples into our tasks and excluded workers who did not pass them.  

We trained seven methods on three different datasets. These methods include six part-prototype based classification methods (ProtoPNet\cite{chen2019looks}, Deformable ProtoPNet\cite{donnelly2022deformable}, SPARROW\cite{kraft2021sparrow}, ProtoTree\cite{nauta2021neural}, TesNet\cite{wang2021interpretable}, and ProtoPool\cite{rymarczyk2022interpretable}) and one unsupervised method (ACE\cite{ghorbani2019towards}). The unsupervised method (ACE) is normally used to extract concepts from a dataset. Each concept is a part of a training sample and therefore is analogous to the concept of prototypes in part-prototype-based methods. As a result, we decided to include it in our experiments as a method representing unsupervised approaches to this problem and also as a baseline.

The datasets on which we trained were CUB-200\cite{WahCUB_200_2011}, Stanford Cars\cite{6755945}, and a subset of the Imagenet dataset\cite{deng2009imagenet} consisting of 7 classes of different felines and canines. The details for our Imagenet subset are included in the appendix. All datasets were augmented in the same process described in the work by Chen et al\cite{chen2019looks}. We also used the public code repositories from the authors of the papers of those methods in all cases except SPARROW. In the case of SPARROW, the authors kindly agreed to give us access to their non-public codebase. We used the same hyperparameters provided by these codebases, only changing those values when needed for different datasets. In particular, we used 10 prototypes for each class when training on datasets where the original codebase did not contain hyperparameters. In the special case of ProtoPool, we used 12 prototypes in total for the ImageNet dataset. We trained all classification methods until they reached at least 70\% top-1 classification accuracy on the dataset. For Imagenet, our criterion was higher at 90\% due to the lower number of classes and the fact that these backbone networks were originally trained on ImageNet in the first place. All methods used ResNet-18 as their backbone network. Something to note is that ProtoPool failed to reach the 70\% accuracy level for the Cars dataset, instead only reaching a Top-1 accuracy of 49.53\%. 

The next step was to gather a suite of query samples from the dataset and generate the explanations given by the methods for classifying those samples. For CUB and Cars, we used the first image found in the directory of each separate class. For Imagenet, we used the first 15 images in order to get enough samples for later use.

One thing we noticed at this stage was that the usual heatmaps provided by many of these methods to denote the location of a prototype or its activation on a query image tended to occlude a large portion of the image itself. Some, like ProtoPNet, tended to draw a rectangle to denote the general boundary of the prototype or its activation. We found this approach to be quite misleading because there were cases where the rectangles could cover more area than the actual activation site. To address this, we opted for a more fine-grained approach where we created boundaries around the area of the activation heatmap with activation values above 70\%. Figure \ref{fig:badrepr} illustrates examples of complex activation sites, failures of the original ProtoPNet region selection approach, and our approach to address the problem.

\section*{Prototype Interpretability}

Our first set of experiments had the goal of finding out the interpretability of the prototypes created by each method. Many part prototype methods claim that their prototypes are intrinsically interpretable. In that case, humans should be able to understand the concept that is encoded within those prototypes. If a prototype represents a large area of an image without any coherent concept understandable to humans, it is not an interpretable prototype.

\begin{figure}[ht]
\centering
\includegraphics[width=0.95\linewidth]{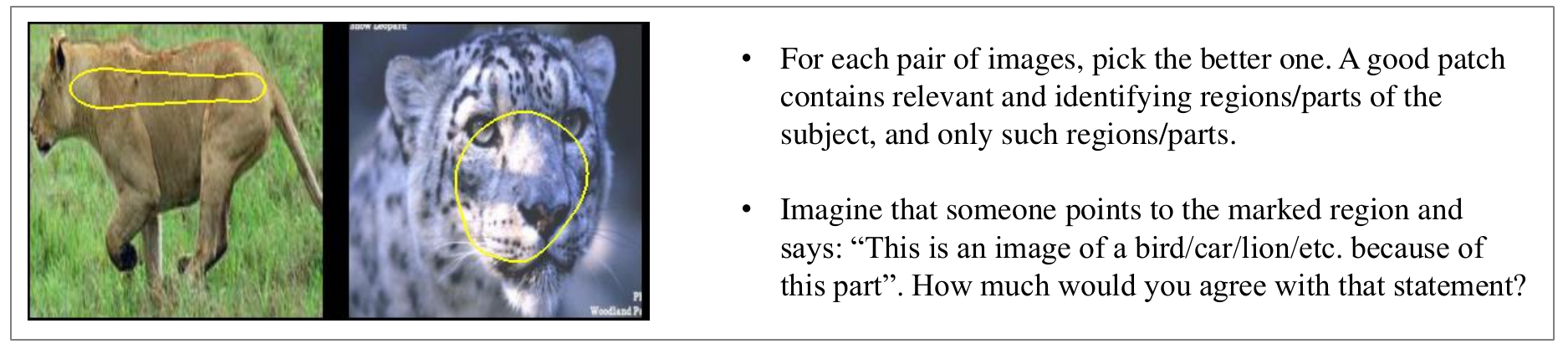}
\caption{Examples of a prototype interpretability comparison task.}
\label{fig:task1}
\end{figure}

In the first experiment, each person was shown a pair of prototypes (concept images in the case of ACE) on the same dataset coming from different methods. The annotators had to choose the prototype they found more interpretable out of the two. Every person had to complete 10 such pairs each time. Out of those 10, 2 were validation pairs handpicked by us to ensure answer quality. An example of such a pair and the instructions given to each person are show in Figure \ref{fig:task1}.


Only people who had finished a qualification task consisting of 7 such pairs with perfect scores were allowed into the main study. Moreover, each 10-pair set of query samples was given to three different people to gauge overall agreement and find individuals who might have chosen randomly. Additionally, this experiment was repeated 3 times and the results were aggregated.

This experiment was a purely comparative study between different methods. This ensured that people had a frame of reference when deciding which method worked better, and this is a common practice in many studies of subjective nature. But they are not without problems. One of the more important problems with this experiment was that it could not answer whether the prototypes from these methods were interpretable or not. It could only answer whether the prototypes of a particular method were \emph{more interpretable on average} than the ones from another. 

To remedy this, we conducted another set of experiments. In the new absolute experiments the annotators were given the option to also choose "Both" and "None" if they thought that both or none of the images in the pair were interpretable. The reason we did not simply show them a single image asking for interpretability was to keep the layout of the experiments very similar. This not only simplified the logistics of creating query samples for the new experiment, but also gave us the opportunity to see whether the rankings of the methods seen in the comparative experiments remained the same when the new options were added. The assumption was that if a very similar experiment with those options added resulted in a drastic change in the rankings, it would reveal problems in our methodology that were hidden beforehand. Instructions given to each person were as follows:

\begin{itemize}
\item For each pair of images, pick the one with the best highlighted patch. A good patch contains relevant and identifying regions/parts of the subject, and only such regions/parts.
\item If both images have good or bad patches, select "Both" or "None" respectively.
\item Imagine that someone points to the marked region and says: "This is an image of a bird/car/lion/etc. because of this part“. How much would you agree with that statement?
\end{itemize}

It is important to note that new image pairs were created for this experiment. This was again to see if a new set of data changes the rankings considerably, revealing any potential hidden problems with the statistical significance of the results. A new qualification task and a validation set were also created. The rest of the setup was almost identical to the comparative experiment. Each time, the annotator was given 10 prototype image pairs and told to pick the best option for each pair, with two out of those 10 pairs coming from the validation set. The experiment was repeated 4 times and the results were aggregated.

\begin{table}[ht]
\centering
\begin{tabular}{|l|l|l|l|l|l|l|l|}
\hline
 & ProtoPNet & Deform. PPNet & SPARROW & ACE & ProtoTree & TesNet & ProtoPool \\
\hline
Total Percentage Selected & 51.65\% & 38.04\% & 38.69\% & 37.64\% & 63.64\% & \textbf{73.26\%} & 47.41\% \\
\hline
\end{tabular}
\caption{Aggregated results for the comparative prototype interpretability experiment}
\label{table1}
\end{table}

\begin{table}[ht]
\centering
\begin{tabular}{|l|l|l|l|l|l|l|l|}
\hline
 & ProtoPNet & Deform. PPNet & SPARROW & ACE & ProtoTree & TesNet & ProtoPool \\
\hline
ProtoPNet & - & 68.93\% & 55.00\% & 66.07\% & 32.97\% & 30.48\% & 53.68\% \\
\hline
Deform. PPNet & 31.07\% & - & 53.57\% & 47.75\% & 30.28\% & 11.58\% & 48.36\% \\
\hline
SPARROW & 45.00\% & 46.43\% & - & 47.87\% & 37.00\% & 18.25\% & 41.73\% \\
\hline
ACE & 33.93\% & 52.25\% & 52.13\% & - & 21.24\% & 24.44\% & 42.86\% \\
\hline
ProtoTree & 67.03\% & 69.72\% & 63.00\% & 78.76\% & - & 43.41\% & 63.54\% \\
\hline
TesNet & 69.52\% & 88.42\% & 81.75\% & 75.56\% & 56.59\% & - & 71.26\% \\
\hline
ProtoPool & 46.32\% & 51.64\% & 58.27\% & 57.14\% & 36.46\% & 28.74\% & - \\
\hline
\end{tabular}
\caption{Results for the comparative prototype interpretability experiment}
\label{table2}
\end{table}

In total, 1104 pairs were compared and annotated by humans for the comparative study. Table \ref{table1} shows the percentage of times each method was chosen against others. As can be seen, prototypes selected by TesNet were chosen 73.26\% of the time when compared to prototypes selected by other methods. In the second place was ProtoTree with 63.64\% preference. It is also interesting to note the relative weakness of SPARROW and Deformable ProtoPNet.

Table \ref{table2} shows the results of the same experiments but on a method-by-method basis. Each element of the table shows the results of the pairwise comparison between the two methods, with the method shown in the first column being the one selected out of the two. For example, the intersection between the row "ACE" and the column "ProtoPool" shows the percentage of times (42.86\%) ACE was chosen against ProtoPool in the experiments. While there is more variation when looked at with this much detail, the general trends hold true.

\begin{table}[ht]
\centering
\begin{tabular}{|l|l|l|l|l|l|l|l|}
\hline
 & ProtoPNet & Deform. PPNet & SPARROW & ACE & ProtoTree & TesNet & ProtoPool \\
\hline
Total Percentage Selected & 65.50\% & 60.26\% & 51.79\% & 53.25\% & 89.56\% & \textbf{95.18\%} & 61.11\% \\
\hline
\end{tabular}
\caption{Aggregated results for the absolute prototype interpretability experiment}
\label{table3}
\end{table}

\begin{table}[ht]
\centering
\begin{tabular}{|l|l|l|l|l|l|l|l|}
\hline
 & ProtoPNet & Deform. PPNet & SPARROW & ACE & ProtoTree & TesNet & ProtoPool \\
\hline
ProtoPNet & - & 55.36\% & 67.74\% & 66.67\% & 62.50\% & 74.58\% & 65.98\% \\
\hline
Deform. PPNet & 53.57\% & - & 74.71\% & 37.04\% & 64.86\% & 57.41\% & 63.33\% \\
\hline
SPARROW & 38.71\% & 49.43\% & - & 37.50\% & 50.65\% & 64.13\% & 64.15\% \\
\hline
ACE & 46.3\% & 74.07\% & 45.83\% & - & 53.23\% & 55.81\% & 42.55\% \\
\hline
ProtoTree & 73.61\% & 89.19\% & 92.21\% & 96.77\% & - & 92.00\% & 97.92\% \\
\hline
TesNet & 89.83\% & 94.44\% & 95.65\% & 95.35\% & 98.00\% & - & 98.18\% \\
\hline
ProtoPool & 55.67\% & 50.00\% & 56.60\% & 68.09\% & 75.00\% & 69.09\% & - \\
\hline
\end{tabular}
\caption{Results for the absolute prototype interpretability experiment}
\label{table4}
\end{table}

For the absolute experiments, 1304 pairs were compared and annotated by humans in total. Table \ref{table3} shows the percentage of times each method's prototypes were chosen in general. As the options for "Both" and "None" also existed in this experiment, the results here show the percentage of prototypes from each method that the annotators found acceptable. While the overall rankings seem to have mostly remained the same, the higher percentages overall show that while some methods might lose in the comparative study, their prototypes are still interpretable, even if a bit less so than the more successful methods. These results also show that 95.18\% of the prototypes created by TesNet and 89.56\% of the ones by ProtoTree are seen as interpretable. These results match our own observations, where it seems that TesNet and ProtoTree offer consistently good prototypes in all of the datasets.

Like before, Table \ref{table4} shows the results of the same experiments but on a method-by-method basis. Each row shows the percentage of times a method (shown on the first column) was selected either solely or as part of the "Both" option when paired with another method shown on the top row. Note that this time, the sum of mirrored cells does not necessarily add up to 100\% due to the existence of "Both" and "None" options. The results show a fine-grained consistency in prototype interpretability rates across all methods.

\section*{Prototype-query Similarity}

\begin{figure}[ht]
\centering
\includegraphics[width=\linewidth]{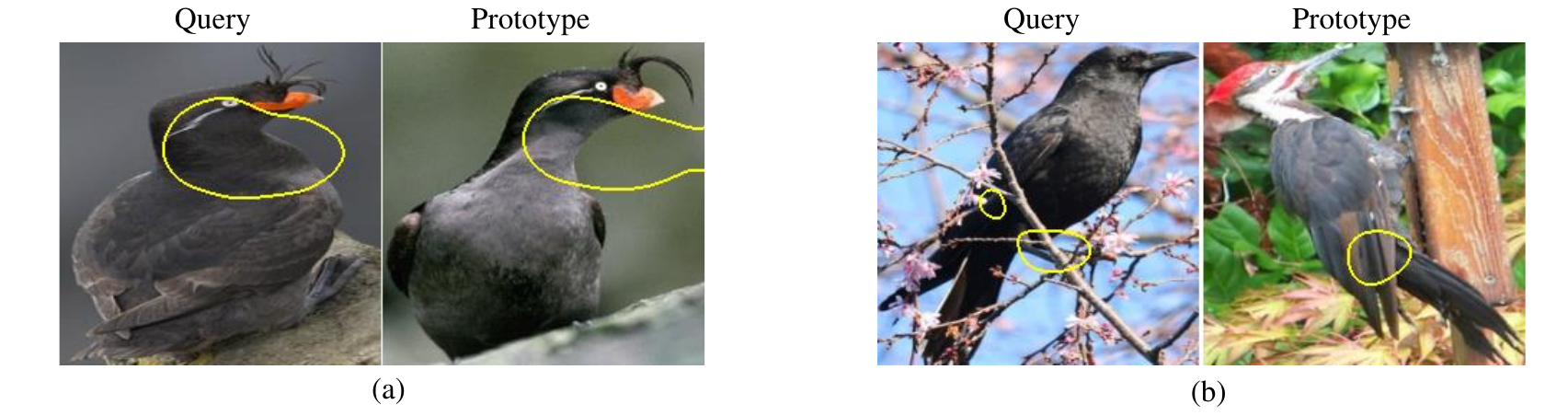}
\caption{ a) An example of a good prototype-query patch similarity. b) an example of a pair of prototype and query patch that are similar in the embedding space but visually very dissimilar.}
\label{fig:badsim}
\end{figure}

\begin{figure}[ht]
\centering
\includegraphics[width=\linewidth]{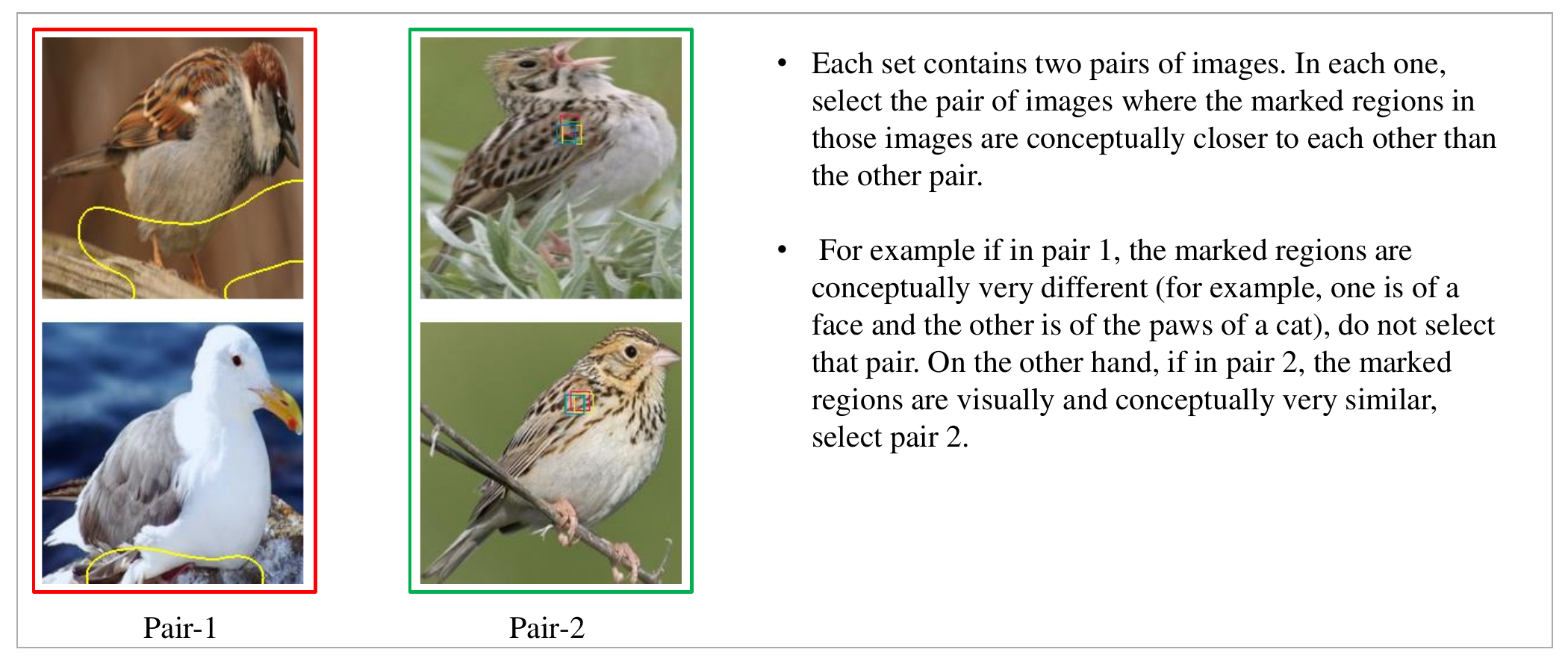}
\caption{ Two pairs of prototype and activated query patch. Annotators were told to pick the most similar pair. In this example, annotators were expected to pick Pair-2 over Pair-1. }
\label{fig:task2}
\end{figure}

Our second set of experiments has the goal of finding out whether the regions activated by a prototype contain the same concept as the prototype itself. Prototype classifiers, in general, use the closest prototype to a query in their operation. Part-prototype classifiers follow the same idea by finding the closest prototype patch to another patch in the input image. For this to be interpretable, the notion of \emph{closeness} used by the method should match that of a human. Otherwise, the explanation itself will seem nonsensical. Figure \ref{fig:badsim} shows examples of both good and bad similarity between the query patch and the prototype.

As before, the first experiment was purely comparative. Each fine-grained task consisted of two pairs of images. Each pair consisted of a prototype and its activation on a query image. The prototypes picked were the most important prototypes when trying to classify an image. The idea was to avoid prototypes that were not instrumental in the final decision as they were more likely to not be present in the query image itself. The annotators were then told to select the prototype/activation pair that were conceptually more similar to each other. Figure \ref{fig:task2} shows examples of this task where annotators are expected to select pair-2 because the corresponding images are more similar than the pair-1 images.

This works well for the part-prototype classification methods as they all have an explanation generation procedure that utilizes their prototypes. On the other hand, ACE is not a classifier. It is an unsupervised concept learning method. Nevertheless, we thought that ACE by itself also had a notion of \emph{similarity} baked into the method. It picks multiple patches from potentially different images to represent a concept. The assumption is that these patches are \emph{ similar} because they share that concept. Here, we decided to simply include two patches from the same concept in the pairs that represented ACE.

The annotators were presented with batches of 10 fine-grained tasks. Like before, 2 out of those 10 tasks contained validation data. Also similar to previous experiments, annotators had to obtain qualification by completing 7 tasks perfectly before being allowed to work on the main experiment. The experiment was repeated 3 times and the results were aggregated. The instructions shown to the annotators are shown in Figure \ref{fig:task2}.


As with prototype interpretability experiments, we also devised a more absolute version of the similarity experiment by adding the options "Both" and "None" to the same experimental layout. The instructions for this other experiment were as follows:

\begin{itemize}
\item Each set contains two pairs of images. In each one, select the pair of images where the marked regions in those images are conceptually closer to each other than the other pair.
\item For example if in pair 1, the marked regions are conceptually very different (for example, one is of a face and the other is of the paws of a cat), do not select that pair. On the other hand, if in pair 2, the marked regions are visually and conceptually very similar, select pair 2.
\item If both pairs have similar regions, select Both. If none of the pairs has similar regions, select None.
\end{itemize}

\begin{table}[ht]
\centering
\begin{tabular}{|l|l|l|l|l|l|l|l|}
\hline
 & ProtoPNet & Deform. PPNet & SPARROW & ACE & ProtoTree & TesNet & ProtoPool \\
\hline
Total Percentage Selected & 64.47\% & 46.12\% & 43.89\% & 22.63\% & 51.93\% & \textbf{71.88\%} & 48.25\% \\
\hline
\end{tabular}
\caption{Aggregated results for the comparative prototype-query similarity experiment}
\label{table5}
\end{table}

\begin{table}[ht]
\centering
\begin{tabular}{|l|l|l|l|l|l|l|l|}
\hline
 & ProtoPNet & Deform. PPNet & SPARROW & ACE & ProtoTree & TesNet & ProtoPool \\
\hline
ProtoPNet & - & 60.17\% & 67.03\% & 87.30\% & 56.80\% & 50.38\% & 66.12\% \\
\hline
Deform. PPNet & 39.83\% & - & 48.48\% & 69.70\% & 46.00\% & 24.79\% & 51.61\% \\
\hline
SPARROW & 32.97\% & 51.52\% & - & 66.34\% & 41.67\% & 19.77\% & 46.88\% \\
\hline
ACE & 12.70\% & 30.30\% & 33.66\% & - & 23.74\% & 13.00\% & 24.17\% \\
\hline
ProtoTree & 43.20\% & 54.00\% & 58.33\% & 76.26\% & - & 33.04\% & 40.23\% \\
\hline
TesNet & 49.62\% & 75.21\% & 80.23\% & 87.00\% & 66.96\% & - & 79.82\% \\
\hline
ProtoPool & 33.88\% & 48.39\% & 53.12\% & 75.83\% & 59.77\% & 20.18\% & - \\
\hline
\end{tabular}
\caption{Results for the comparative prototype-query similarity experiment}
\label{table6}
\end{table}

In total, 1156 pairs were compared and annotated by humans for the comparative study of prototype-query similarity. Table \ref{table5} shows the percentage each method was chosen against others. As can be seen, TesNet were chosen 71.88\% of times when compared to other methods. At the second place was ProtoPNet with 64.47\% preference. This shows that apart from TesNet, the other methods show comparatively worse prototype-query similarity compared to ProtoPNet itself. This has an interesting implication for the interpretability of these methods compared to the baseline.

\begin{table}[ht]
\centering
\begin{tabular}{|l|l|l|l|l|l|l|l|}
\hline
 & ProtoPNet & Deform. PPNet & SPARROW & ACE & ProtoTree & TesNet & ProtoPool \\
\hline
Total Percentage Selected & 40.41\% & 35.03\% & 36.74\% & 30.56\% & 34.97\% & \textbf{42.98\%} & 31.71\% \\
\hline
\end{tabular}
\caption{Aggregated results for the absolute prototype-query similarity experiment}
\label{table7}
\end{table}

Table \ref{table6} shows the results of the same experiments but on a method-by-method basis. Each element of the table shows the results of the pairwise comparison between the two methods, with the method shown in the first column being the one selected out of the two. For example, the intersection between the row "ACE" and the column "ProtoPool" shows the percentage of times (24.17\%) ACE was chosen against ProtoPool in the experiments. While there is more variation when looked at with this much detail, the general trends hold true.

For the absolute experiments, 1984 pairs were compared and annotated by humans in total. Table \ref{table7} shows the percentage of times each method's prototypes and query sample activations were seen to be close conceptually. While the overall rankings seem to have mostly remained the same, the lower overall percentages show that none of these methods is doing well in this regard. The best results, TesNet at 42.98\% and ProtoPNet at 40.41\% are still quite low. This is in agreement with the results previously reported in \cite{hoffmann2021looks, kim2022hive}. 

\begin{table}[ht]
\centering
\begin{tabular}{|l|l|l|l|l|l|l|l|}
\hline
 & ProtoPNet & Deform. PPNet & SPARROW & ACE & ProtoTree & TesNet & ProtoPool \\
\hline
ProtoPNet & - & 39.56\% & 35.29\% & 31.18\% & 46.23\% & 49.02\% & 37.66\% \\
\hline
Deform. PPNet & 35.16\% & - & 28.07\% & 28.92\% & 33.98\% & 52.53\% & 32.20\% \\
\hline
SPARROW & 30.88\% & 47.37\% & - & 27.62\% & 30.67\% & 35.35\% & 43.10\% \\
\hline
ACE & 29.03\% & 31.33\% & 33.33\% & - & 31.71\% & 25.93\% & 31.19\% \\
\hline
ProtoTree & 36.79\% & 32.04\% & 34.67\% & 29.27\% & - & 40.37\% & 35.14\% \\
\hline
TesNet & 44.12\% & 46.46\% & 31.31\% & 35.80\% & 53.21\% & - & 45.00\% \\
\hline
ProtoPool & 33.77\% & 27.97\% & 32.76\% & 26.61\% & 43.24\% & 30.00\% & - \\
\hline
\end{tabular}
\caption{Results for the absolute prototype-query similarity experiment}
\label{table8}
\end{table}

Finally, Table \ref{table8} shows the results of the same experiments but on a method-by-method basis. Each row shows the percentage of times a method (shown on the first column) was selected either solely or as part of the "Both" option when paired with another method shown on the top row. Note that the sum of mirrored cells does not necessarily add up to 100\% due to the existence of "Both" and "None" options. The consistently low similarity rates across different methods point to a general failure of all models in achieving a good similarity between the activated regions of the query samples and the prototypes.

\section*{Interpretability of the Decision-Making Process}

Our final experiment was designed to assess the interpretability of the decision-making process. In general, part-prototype methods offer a set of explanations for a decision and then use those concepts for a final classification step, which is usually a fully connected layer. We designed our framework for cases where the final decision is made using such a final layer over all of the prototype activations and so we tried to stay as faithful as possible to the actual decision-making process. This meant that we had to make sure we included the activation scores and rankings of the prototypes in the experiment design. 

ACE is not a classification method and there is no final decision to make. As a result, ACE, alongside ProtoTree, is not included in this experiment. In the case of ProtoTree, the method uses a decision-tree structure to offer its explanations. This is not compatible with our experiment design. Moreover, the agreement and distinction tests designed by HIVE\cite{kim2022hive} are adequate for evaluating the interpretability here as they use the same tree structure as the method itself. In the end, the authors of HIVE found that the tree structure was interpretable by humans. This cannot be said for the remaining five methods. 

\begin{figure}[t]
\centering
\includegraphics[width=0.45\linewidth]{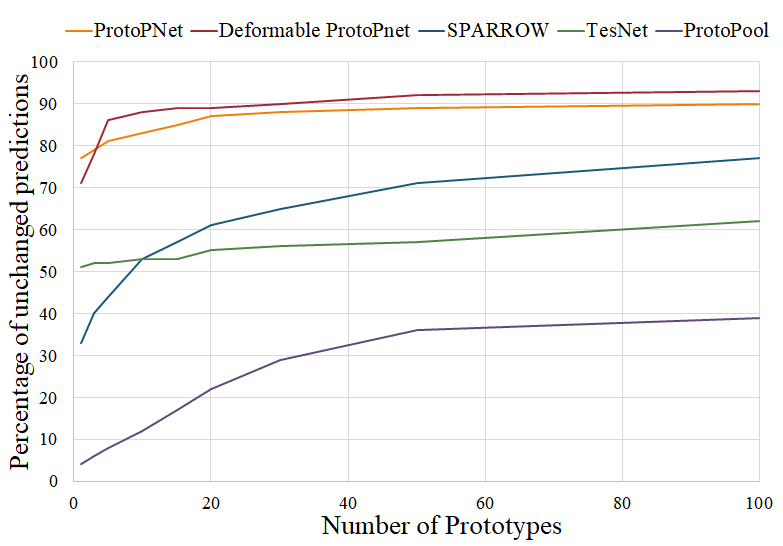}
\caption{The degree to which a reduced model makes decisions similar to the model that uses the complete prototype set. The horizontal axis shows the number of prototypes retained in a reduced model. The vertical axis is the percentage of test data where the predictions made by the reduced model remain the same as the prediction made by the model that uses the complete prototype set on the CUB dataset.}
\label{fig:methodprotacc}
\end{figure}

A challenge in interpretability is to show enough information to humans so that they can understand the process without overwhelming them. For example, we cannot expect a human to look at the entire 2000 prototypes and their activations in order to understand how ProtoPNet classifies a query image from the CUB dataset. Many part-prototype models seem to have opted to use the 10 most activated prototypes for a particular query image as a good proxy for the entire model\cite{donnelly2022deformable, chen2019looks, wang2021interpretable}. 10 prototypes are much more manageable for a human to understand than 2000, but we have to make sure that they are also a good proxy for the model itself. 

If the model, using those same 10 prototypes, is able to classify the query image and come to the same decision as before (regardless of whether that original decision was correct or not), then those 10 prototypes are a good proxy for the decision-making process of the model for that particular query image. This implies that the model selectively utilizes a limited number of prototypes from its extensive pool to categorize images.

This is unfortunately not always the case. For some query images, the model might use more than 10 prototypes to come up with its final decision. In those cases, we can argue that the interpretability of the method will suffer, as it relies on a higher number of prototypes, which, in turn, can overwhelm humans. As a result, finding out the model's accuracy with only a limited number of prototypes can be a good indicator of the interpretability of the decision-making process. We performed experiments where we only kept the top 1, 3, 5, 10, 15, 20, 30, 50, and top 100 prototypes for each method on the CUB dataset, and For each reduced model, we calculated the percentage of its decisions that agree with the decisions of the model that uses the complete prototype set. The results of these can be seen in Figure \ref{fig:methodprotacc}.

As can be seen, not all methods are equal in this regard. Some like ProtoPNet and Deformable ProtoPNet can achieve the same decisions using relatively few prototypes. Others like ProtoPool utilize a large amount of prototypes to classify most of the query images. Only 12\% of the decisions of ProtoPool remain the same when only the 10 top prototypes are kept. This observed low percentage is concerning especially as it only goes up to 39\% after using 100 prototypes out of the 202 prototypes this model had for the CUB dataset.

\begin{figure}[ht]
\centering
\includegraphics[width=\linewidth/2]{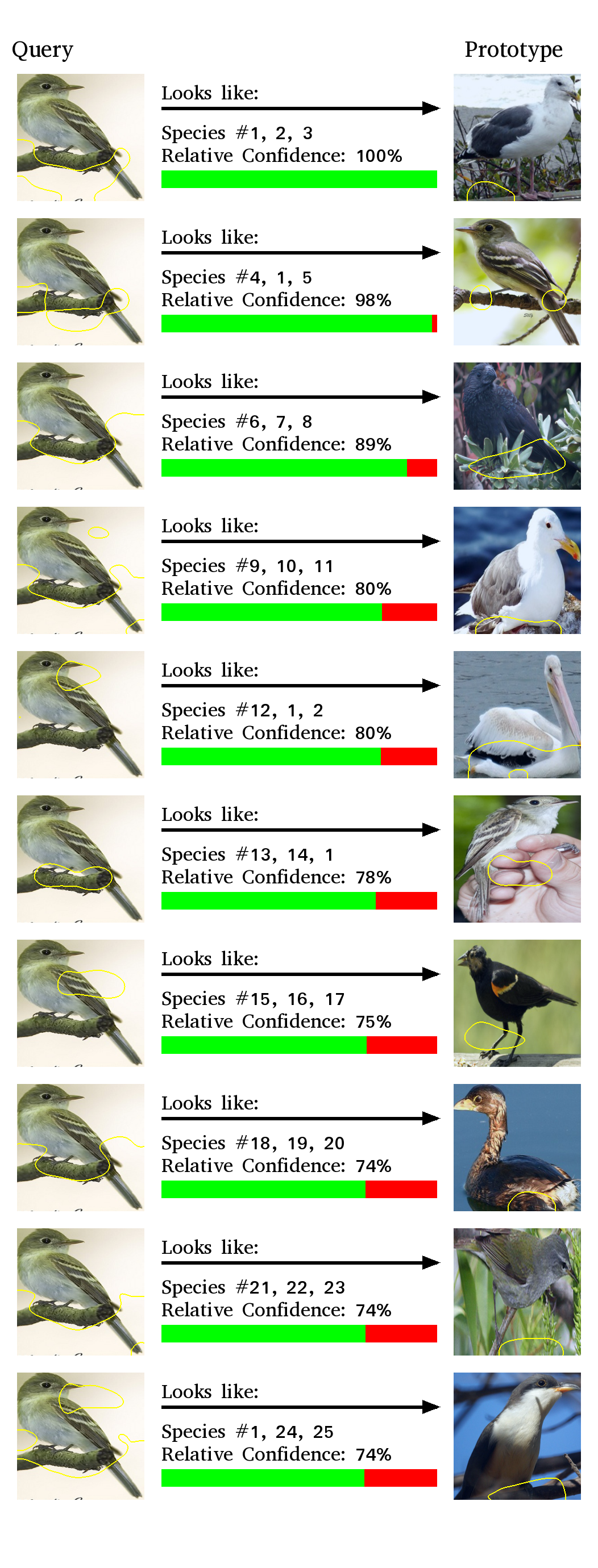}
\caption{Example of an Explanation set used in the experiment about the interpretability of the decision-making process itself.}
\label{fig:task3example}
\end{figure}

After this step, we classified the same query samples utilized in the previous experiments using our models but only kept the ones where the decision of the model did not change after keeping only the 10 top activated prototypes. For each of these query images that passed this filter, we then gathered the top 10 prototypes and their activation regions on the query image into an explanation set. The prototypes in this explanation set were sorted from the highest activation to the lowest. As the activation values themselves might be different for each query image and might not mean anything to a human, we used a confidence system where the top activated prototype has a relative confidence level of 100\% and every other prototype has its confidence level scaled by its activation value compared to the activation value of the top prototype. We then included the class label associated with that prototype in the explanation set. In the case of ProtoPool which doesn't have a single class associated with each prototype, we picked the 3 top classes associated with that prototype. These classes were obfuscated with numbers so that the role of prior human knowledge could be reduced. Examples of these explanation sets can be found in Figure \ref{fig:task3example}. We then asked our human participants on Amazon Mechanical Turk to guess the class that was picked for that query image by our classifier model. Our instructions for the AMT participants were as follows:

\begin{itemize}
    \item Given a photo of an animal or a car (Query), an AI is predicting the species or the model of the subject of the photo using prototypes it has learned from previously seen photos. Specifically for each prototype, the AI identifies a region in the photo (Query) that looks the most similar to a region (marked with a yellow border) in the prototype and rates its relative confidence in their similarity. Each prototype is associated with one, or multiple animal species/car models. The AI will pick the species/type based on how similar and frequent the prototypes are to the query.
    \item For each query, we show explanations on how the AI reasons the species of the animals or the model of the car. Looking at the species/model numbers for each query-prototype pair, guess the species/model number that you would think the AI predicted for the query.
    \item Guessing randomly will get you low overall accuracy depending on the number of options. You will only get rewards if your performance is sufficiently higher than random guesses.
    \item Remember that you can zoom-in/out and pan the image using the buttons below the image.
\end{itemize}

The idea behind this experiment is similar to that of the distinction task from HIVE. In order for a human to understand the decision-making process of an AI system, they should be able to correctly predict the decision of the AI system from the same information it uses. 



\begin{table}[ht]
\centering
\begin{tabular}{|l|l|l|l|l|l|l|l|}
\hline
 & ProtoPNet & Deform. PPNet & SPARROW & TesNet & ProtoPool \\
\hline
Percentage of Correct Predictions & 96.01\% & 89.92\% & 85.40\% & 92.86\% & 48.77\% \\
\hline
\end{tabular}
\caption{Results for the interpretability of the decision-making process. Numbers are the percentage of model decisions that human participants could correctly predict based on the explanation given by the model.}
\label{table9}
\end{table}

In total, 607 explanation sets were shown to and annotated by the participants. Table \ref{table9} shows the percentage of query samples in which humans correctly predicted the predictions of the models. Apart from ProtoPool, human participants seem to be able to understand the underlying principles behind the final decision-making process of the rest of the methods. Prototypes in ProtoPool can belong to multiple classes and this causes severe issues with understanding how the method makes its decisions. The stark contrast between ProtoPool and the others shows that this evaluation criteria is a useful way of determining problems in the interpretability of Prototype-based methods.

\subsection*{The Effects of Low Prototype Counts}

During preparation for our experiments, we noticed an unexpected phenomenon for prototype similarity. In some dataset/method combinations, it was possible to achieve high classification accuracies of more than 80\% with only one prototype per class. For example, TesNet was able to achieve about 88\% accuracy when trained on the ImageNet subset using only 7 prototypes. In this instance, the similarity of a prototype with its activation was very low. This sheds light on some of the limitations of these methods. 

Part-prototype-based classifiers are, in the end, discriminative models. This is in contrast to how they are perceived from an interpretability standpoint. When humans look at the explanations given for the decision, they see the similarity between the prototypes and the activated regions. However such similarity is only useful to the model itself as long as it can utilize it to discriminate effectively between the classes. As long as a query sample is closer to the prototype of the correct class than all other classes, the model makes the correct decision, regardless of the actual similarity between the query sample and the prototype. When designing a model, setting the number of prototypes to a small value favors interpretability, but, on the other hand, it may adversely affect the similarity of the prototypes and their corresponding activations. 
Even when the number of prototypes is relatively small, to truly understand the model, it is essential to examine the similarity between prototypes and their corresponding activations (as in experiment 2), as well as the interpretability of the individual prototypes (as in experiment 1).




 

\section*{Discussions}

In a substantial portion of the previous work, it was implicitly assumed that part-prototype networks are interpretable by nature. We have devised three human-centric evaluation schemes to asses that assumption. One for evaluating the interpretability of the prototypes themselves, one for evaluating the similarity between prototypes and the activated regions of the query sample, and one for evaluating the interpretability of the decision-making process itself.  Our experiments show that these schemes are able to differentiate between various methods in terms of interpretability while not suffering from the problems of the previous works. Moreover, we applied this scheme to seven related methods on three datasets. The results shed light on the interpretability of these methods from a human perspective.

The results show that not all part-prototype methods are equal when it comes to human interpretability. In some cases, there are severe issues that hurt the interpretability of these methods. Chief among them is the dissimilarity between the prototype and the activation region of the query. This problem existed for all of the models tested to some degree. In addition, the prototypes of some models were not sufficiently interpretable. Finally, ProtoPool has a noticeable problem with the interpretability of the decision-making process due to assigning multiple classes to each prototype and requiring huge numbers of prototypes to make their final decisions. The decision-making process of some other methods were easy to understand using the top 10 activated prototypes, but they still needed more than 10 prototypes to make their final decision.

Still, in some other cases, the methods performed relatively well. Apart from ProtoPool, all methods performed quite well in the interpretability of the decision-making process test. ProtoPNet and Deformable ProtoPNet were also able to make the majority of their predictions using the top 10 activated prototypes. Prototype interpretability was also particularly high for some of the methods involved. Considering these results and also the observations on low prototype count models, our suggestion is to look at the top prototype sets for the final decision rather than only individual prototypes to get a better understanding of a model. Even if the most activated prototype has low interpretability or is dissimilar to its activation, the other top prototypes used in the decision could better explain the final decision. It is important to note that the opposite could also happen: the top prototype being interpretable but the others showing a flawed decision-making process. 

The interpretability of many machine learning methods used in production is very important. However, it is more important to understand the limitations and peculiarities of the models, methods, and tools used to address the interpretability problem. In particular, unified frameworks with emphasis on human-centric assessments are necessary if we want to truly evaluate interpretability methods. We think that more studies like this should be done on different areas of AI interpretability to further our understanding of such models.

\section*{Author Contributions}

O.D. Designed, developed, and implemented the experiments as well as wrote the paper, S.M. developed and implemented the experiments, and M.K. designed the experiments and wrote the paper.

\section*{Data availability}

Anonymized data of the experiments can be found in the following repository: https://github.com/omiddavoodi/part-prototype-interpretability-data

\section*{Competing Interests}

The authors declare no competing interests.

\bibliography{sample}

\end{document}